\documentclass[sigconf]{acmart}
\usepackage{tikz}
\usepackage{standalone}

\usepackage{booktabs}
\usepackage{multirow}
\usepackage{makecell}
\usepackage{algorithm, algorithmic}
\usepackage[most]{tcolorbox}
\setcopyright{acmlicensed}
\copyrightyear{2025}
\acmYear{2025}
\acmDOI{}
\acmConference[TGL Workshop, KDD]{KDD}{2025}{Toronto, ON}

\acmISBN{}

\title{Base3: a simple interpolation-based ensemble method for robust dynamic link prediction}

\newcommand{\first}[1]{\textbf{\textcolor{red}{#1}}}
\newcommand{\second}[1]{\underline{\textcolor{blue}{#1}}}
\newcommand{\third}[1]{\emph{\textcolor{violet}{#1}}}

\begin{document}

\author{Emma Kondrup}
\email{emma.kondrup@mila.quebec}
\orcid{0009-0007-3480-302X}

\renewcommand{\shortauthors}{Kondrup}

\begin{abstract}
    Dynamic link prediction remains a central challenge in temporal graph learning, particularly in designing models that are both interpretable and effective. Existing approaches often rely on complex neural architectures, which are computationally intensive and difficult to interpret. 
    In this work, we build on the strong recurrence-based foundation of the EdgeBank baseline \cite{edgebank}, by supplementing it with inductive capabilities. We do so by leveraging the predictive power of non-learnable signals from two perspectives that complement EdgeBank's historical edge recurrence: 
    global node popularity, as introduced in the PopTrack \cite{PopTrack} model, and co-occurrence patterns through our proposed module, \textbf{t-CoMem}. t-CoMem is a lightweight memory module that tracks temporal co-occurrence patterns and neighborhood activity. Building on this, we introduce \textbf{Base3}, an interpolation-based model that fuses EdgeBank, PopTrack, and t-CoMem into a unified scoring framework. This combination effectively bridges local and global temporal dynamics -- repetition, popularity, and context -- without relying on training. Evaluated on the Temporal Graph Benchmark, Base3 achieves performance competitive with state-of-the-art deep models, even outperforming them on some datasets. Importantly, it considerably improves on existing baselines' performance under more realistic and challenging negative sampling strategies -- offering a simple yet robust alternative for temporal graph learning. 

    \vskip .1in 

    The code used in this work is available \href{https://github.com/ekmpa/Base3}{\texttt{here}}.

\end{abstract}


\received{May 2025}

\maketitle

\section{Introduction}

Many real-world networks (i.e., social and financial platforms, or communication logs) are dynamic by nature and evolve continuously over time. While static graph-based models have achieved notable success in capturing structural dependencies \cite{kipf2017, velickovic2018}, they fall short in representing the temporal evolution of interactions. This gap has spurred the development of several temporal graph learning methods \cite{kazemi2020representation}, with many tackling graph-based tasks with dynamic information \cite{edgebank}. However, many of these methods rely on deep neural architectures that require extensive message passing, large-scale training, and careful tuning \cite{rossi2020temporal, xu2020inductive, trivedi2019}. Their high computational cost often makes them impractical for real-world deployment, and their complexity can considerably limit interpretability \cite{PopTrack}.

\vskip .1in

In this work, we challenge the notion that such complexity is necessary. We propose lightweight, training-free alternatives that exploit simple yet powerful temporal signals, namely edge recurrence and node popularity. Building on the success of two recent non-learnable baselines -- EdgeBank, which memorizes past edges \cite{edgebank}, and PopTrack, which models temporal popularity \cite{PopTrack} -- we introduce \textbf{t-CoMem}, a memory-based module that captures co-occurrence and neighborhood-level activity over time. We further present \textbf{Base3}, an interpolation-based model that combines EdgeBank, PopTrack, and t-CoMem into a unified scoring framework.

\vskip .1in 

Despite their simplicity, our models perform competitively with state-of-the-art deep learning methods on the Temporal Graph Benchmark (TGB) \cite{TGB}. Notably, they show exceptional robustness across challenging evaluation settings, including historical and inductive negative sampling -- scenarios where existing models often degrade. This demonstrates that carefully designed non-learnable models can offer not only interpretability and efficiency but also strong generalization in realistic dynamic graph tasks.

\section{Related Work} 

\textbf{Heuristic Approaches for Link Prediction.} Before the rise of neural models, link prediction in graphs was typically approached using heuristic-based methods. Some of these remain strong, interpretable baselines today \cite{liben2007link}. These heuristics exploit simple topological signals to estimate the likelihood of a link forming between two nodes. Among the most well-known are the \textit{Common Neighbors} and \textit{Preferential Attachment} measures. While these are limited in performance, especially in complex settings, they are quite straightforward in respect to both not requiring any training and offering interpretable insights.

\vskip .1in

An important principle behind many link prediction heuristic approaches is that of \textit{Triadic Closure}, which suggests that if two nodes share a common neighbor, they are more likely to, themselves, form a direct connection \cite{easley2010networks}. It reflects tendencies toward triangle formation in real-world networks, especially social networks, which often exhibit high levels of triadic closure. 

\vskip .1in

Building on this, the \textit{Common Neighbors} score measures the size of the intersection between the neighbor sets of two nodes:
\begin{equation}
\text{CN}(u, v) = |\Gamma(u) \cap \Gamma(v)|
\end{equation}
where $\Gamma(x)$ denotes the set of neighbors of node $x$. This heuristic is particularly effective in networks where triadic closure is common \cite{liben2007link}.
Another widely used measure is \textit{Preferential Attachment}, which is grounded in generative network theory. It assumes that high-degree nodes are more likely to form new links—a phenomenon often described as ``the rich get richer''. Its formulation is:
\begin{equation}
\text{PA}(u, v) = |\Gamma(u)| \cdot |\Gamma(v)|
\end{equation}
This heuristic is especially relevant in scale-free networks, such as citation graphs or web data \cite{barabasi1999emergence}. Along with the \textit{Adamic Adar Index}, defined as 
\begin{equation}
    AA(u,v) = \sum_{v\in\Gamma(u)\cap \Gamma(v)} \frac1{\log | \Gamma(w)|}
\end{equation} and the \textit{Resource Allocation Index}, defined as 
\begin{equation}
    RA(u,v) = \sum_{w\in \Gamma(u) \cap \Gamma(v)} = \frac1{|\Gamma(w)|} 
\end{equation} these are shown to reach results comparable with state-of-the-art methods, while offering more interpretability \cite{dileo2024link, cornell2025power}. They also offer insight into the structural biases that more complex models aim to learn or surpass \cite{liben2007link, lichtenwalter2010new}.

\vskip .1in 

\textbf{Static Graph Neural Networks (GNNs)} have become foundational tools for learning on relational data. In static graphs, where the node and edge sets remain fixed, GNNs learn by aggregating and transforming features from local neighborhoods \cite{kipf2017}. Their inherent ability to model natural dependencies between entities in the graph makes them particularly effective at capturing local structural patterns. Variants of the original GNN architecture quickly emerged, notably incorporating attention mechanisms \cite{velickovic2018} which allowed the model to learn dynamic weighting of neighbors based on their relative importance and thus moved beyond uniform aggregation. GraphSAGE, another widely adopted variant, introduced neighborhood sampling for inductive representation learning, enabling generalization to unseen nodes in large-scale static graphs \cite{hamilton2017}. 

\vskip .1in 

Despite their usefulness, these models remain assuming a fixed topology, which limits their applicability to real-world domains such as communication networks, transportation or web data, where interactions and relationships are inherently dynamic. These evolving structures call for models that can adapt to temporal changes in the graph, motivating research in dynamic and temporal GNNs.

\vskip .1in 

\textbf{Temporal Graph Learning} focuses on modeling spatial and temporal dependencies in evolving networks. Following the taxonomy in \cite{kazemi2020representation}, methods are broadly divided into Discrete-Time Dynamic Graphs (DTDGs) and Continuous-Time Dynamic Graphs (CTDGs), depending on how they represent temporal evolution.

\vskip .1in

\textit{DTDG methods} represent temporal dynamics using a sequence of graph snapshots sampled at fixed intervals. This discretized approach enables the reuse of static GNNs in conjunction with recurrent or temporal modules to capture historical patterns over time \cite{pareja2020evolvegcn, skarding2021foundations}. While DTDGs offer computational efficiency and a straightforward temporal abstraction, they often struggle to capture fine-grained or asynchronous event dynamics, and may smooth over important temporal details \cite{trivedi2019}. Recent work, such as the recently-proposed Unified Temporal Graph \cite{huang2024utg}, seeks to bridge this gap by adapting snapshot-based models to handle irregular event streams.

\vskip .1in 

In contrast, \textit{CTDG methods} capture interactions at precise timestamps, treating the graph as an asynchronous sequence of interactions. This finer granularity better reflects the irregular, event-driven nature of many real-world systems, such as financial transactions, messaging platforms, or online user behavior \cite{rossi2020temporal}. Notable CTDG models include TGAT, which introduces temporal attention and time encoding \cite{xu2020inductive} and TGN, which incorporates memory modules and message queues for long-term temporal context \cite{rossi2020temporal}. More recent models like GraphMixer \cite{cong2023graphmixer} use parameter-efficient mixing layers to combine structural and temporal signals, achieving strong performance on large-scale dynamic graphs. DyGFormer \cite{yu2023towards} and TNCN \cite{zhang2024efficient} further improve temporal modeling by introducing transformer-based spatiotemporal attention and context-aware temporal convolutions, respectively. Together, these models highlight the importance of explicitly modeling temporal granularity and memory in CTDG frameworks to capture the full complexity of evolving graph data.

\vskip .1in 

Despite their expressive power, however, these usually heavy and complex models require extensive training data, computational capabilities, and careful tuning. 

\vskip .1in 

\textbf{Dynamic Link Prediction} is one of the principal tasks in Temporal Graph Learning, and consists of predicting the existence of an edge between two existing nodes at a given timeframe. The added complexity of dynamic behavior and evolving communities makes this task considerably harder than static link prediction.  

\vskip .1in

\textbf{Negative Sampling Strategies} play a crucial role in the evaluation of temporal learning models, as highlighted in \cite{edgebank}. They directly impact the difficulty of the prediction task and thus the interpretability of performance metrics. The most commonly used method has been \textit{random negative sampling}, where negative edges (non-existent links) are simply uniformly sampled from the set of all possible node pairs \cite{rossi2020temporal, xu2020inductive}.  While efficient, random sampling may yield trivial negatives -- node pairs that are structurally or temporally unrelated -- which can inflate confidence signals and misrepresent a model’s generalization ability \cite{edgebank}. 

\vskip .1in

To address these limitations, recent work has proposed more challenging and realistic alternatives. These alternatives are crucial for confidently validating temporal graph models under realistic settings. The EdgeBank framework \cite{edgebank}, which marked an important step in highlighting these limitations, formalizes two such strategies: \textit{inductive} and \textit{historical} sampling. Inductive sampling evaluates a model’s ability to generalize to unseen nodes by constructing negative samples that include entities not observed during training. This setting simulates real-world cold-start or deployment scenarios where models must infer relationships for entirely new entities. Historical sampling, on the other hand, draws negatives from node pairs that have interacted in the past but are not linked in the current prediction window. These "near-positive" negatives are more ambiguous, requiring the model to distinguish between truly inactive and merely temporarily inactive links. Both strategies increase the robustness and credibility of evaluation by focusing on more realistic and informative failure cases, aligning with the broader need for standardized, challenging benchmarks in temporal graph learning \cite{huang2024utg}.

\vskip .1in

\textbf{EdgeBank} and \textbf{PopTrack} represent two of the most simple-yet-competitive non-learnable baselines for dynamic link prediction.

\vskip .1in 
\textit{EdgeBank}, introduced by Poursafaei et al. \cite{edgebank}, is grounded in the principle following which past links are likely to re-occur. It maintains a memory of all previously seen edges—its \textit{edge bank}—and predicts future links by checking whether a candidate edge exists in this memory. EdgeBank has two forms, EdgeBank$_{\text{ tw}}$ which keeps an edge bank over a recent determined time window, and EdgeBank$_{\infty}$ for which the edge bank spans the entire timeframe. This memorization-based approach proves highly effective in domains with strong recurrence patterns. However, EdgeBank is fundamentally non-inductive: it cannot predict links involving node pairs never observed during training, limiting its generalization to novel interactions. Despite this, EdgeBank achieves highly competitive performance, particularly in domains with recurring relationships -- and has thus since become a widely-used baseline in the temporal graph learning literature \cite{TGB, edgebank}. Existing efforts to supplement EdgeBank with inductive capabilities have consisted in incorporating it with temporal collaborative filtering, a method which, while interesting, has yet to show itself to be highly competitive in terms of performance \cite{mohammadzadeh2024temporal}. 

\vskip .1in 
\textit{PopTrack} \cite{PopTrack}, on the other hand, builds on the principle that node popularity correlates with connectivity. It predicts a link from node $u$ to node $v$ if the incident node $v$ ranks among the top-$K$ most popular nodes at time $t$, based on recent interaction frequency. Popularity is tracked using a decayed count of incoming edges, emphasizing recency while retaining longer-term trends. This makes PopTrack particularly well-suited for non-stationary environments, and where temporal bursts or shifting popularity drive link formation more than repeated edge patterns. Unlike EdgeBank, PopTrack can generalize to unseen links as long as the destination node has accumulated sufficient popularity—providing a lightweight but effective form of inductive reasoning.

\vskip .1in 

\section{Methods}

Our proposed \textbf{t-CoMem} module and \textbf{Base3} model build on the inductive biases of the two non-learnable baselines -- EdgeBank and PopTrack -- by embedding them within a simple, interpretable, and training-free framework. This framework is designed with two key goals in mind: (1) to evaluate whether straightforward memory and aggregation mechanisms can match or outperform complex neural architectures, and (2) to offer viable solutions for low-resource or real-time deployment scenarios where efficiency and interpretability are paramount.

\subsection{t-CoMem}

\textbf{t-CoMem (Temporal Co-occurrence Memory)} is a non-parametric module designed to combine two main ideas: temporal co-occurrence tracking and recent popularity weighting.

\vskip .1in 

 t-CoMem maintains a dynamic memory by tracking how frequently node pairs co-occur within a fixed time window (set to 1,000,000 by default), capturing short- to mid-range temporal dependencies through co-appearance patterns. To enrich this signal, it incorporates a soft popularity score from PopTrack, weighting nodes by their recent activity (their popularity score) rather than relying on binary top-$K$ membership. Unlike PopTrack, which considers only the destination node, t-CoMem also factors in the source’s recent interactions—addressing a limitation highlighted in \cite{PopTrack} and promoting more context-aware predictions.

\vskip .1in

\begin{tcolorbox}[
  title=t-CoMem Implementation Details, 
  colback=gray!5!white, 
  colframe=gray!80!black,
  breakable,            
  enhanced,             
  before upper={\parindent15pt},
  before skip=10pt, after skip=10pt  
]
\textbf{Hyper-parameters:}
\begin{enumerate}
    \item Time window \texttt{tw}, determines the wanted recency period to consider;
    \item Co-occurence weight $\lambda$, determines how strongly co-occurence affects scoring. 
\end{enumerate}

\vskip .1in 

\textbf{Data Structures:} 
\begin{enumerate}
    \item A mapping from each node $u$ to a deque of its most recent destination nodes, $\mathcal{D}[u]$, timestamped and bounded by the time window.
    \item A dictionary storing the co-occurrence count for each node pair, $\mathcal{C}[u][v]$. 
\end{enumerate}

\vskip .1in 
\textbf{Memory updates:} memory is built through batch updates. For each batch of 200 edges, for each edge $(u,v,t)$:
\begin{itemize}
    \item Append $(t, v)$ to $\mathcal{D}[u]$
    \item $\mathcal{C}[u][v] \gets \mathcal{C}[u][v] + 1$
    \item $\mathcal{C}[v][u] \gets \mathcal{C}[v][u] + 1$
\end{itemize}

\vskip .1in

\textbf{Scoring:} combines neighborhood popularity and direct co-occurrence. To do so, t-CoMem:

\begin{itemize}
    \item Retrieves all recent destinations from $\mathcal{D}[u]$ within the time window.
    \item For each recent neighbor $n_i$, increment the score by $n_i$'s PopTrack popularity $p$, exponentially decayed by how recently it was observed:
    \begin{equation}
    \text{decayed\_score} = \sum_{n_i} d \cdot p_i
    \end{equation}
    where \( d = \exp\left(-\frac{t - t_i}{\texttt{tw}}\right) \). This decay introduces recency bias, making recommendations more relevant in the common context in which recent interactions carry more predictive power, and ensuring smooth forgetting which aligns with real-world patterns. 

    \vskip .1in
    
    \item Retrieves the co-occurrence count $c$ between $u$ and $v$, and computes its influence $f$ as:
    \begin{equation}
    f = \lambda \cdot \frac{c}{1 + c}
    \end{equation}

    \item Returns the combined score using:
    \begin{equation}
    \text{score}_{\text{ t-CoMem}} = \frac{1}{1 + \frac{1}{\sum d \cdot p + f}}
    \end{equation}
    which squashes the result to the range $[0, 1]$.
\end{itemize}

\end{tcolorbox}

This way, recent activity is given importance while multi-hop propagation can occur through the propagation of PopTrack popularity within these recent neighbour lists.  t-CoMem's design addresses key limitations of purely popularity-based or recurrency-based models like PopTrack or EdgeBank respectively, especially PopTrack's inability to condition predictions on the source node.

\subsection{Base3}

\textbf{Base3} is an ensemble interpolation model that linearly combines the prediction scores from EdgeBank, PopTrack, and t-CoMem. Each component contributes a weighted vote to the final score, offering a hybrid prediction that balances recurrence (EdgeBank), popularity (PopTrack), and co-occurrence (t-CoMem) -- thus fusing complementary inductive signals in a modular fashion. Our proposed Base3 presents itself as a strong model that outperforms a majority of existing models, learnable and non-learnable alike. 

\begin{table*}[ht]
\caption{MRR score comparison with learnable models, where the \first{first}, \second{second} and \third{third} best performances are highlighted.   * denotes our contributions.}
\label{fullMRR}
\centering
\small
\begin{tabular}{lcccccc}
\toprule
\textbf{Method} & \multicolumn{2}{c}{\textbf{tgbl-wiki-v2}} & \multicolumn{2}{c}{\textbf{tgbl-review-v2}} & \multicolumn{2}{c}{\textbf{tgbl-coin}} \\
 & Validation MRR & Test MRR & Validation MRR & Test MRR & Validation MRR & Test MRR \\
\midrule
Base3 * & \second{0.727} & \second{0.743} & 0.101 & 0.108 & \first{0.754} & \first{0.773} \\ 
t-CoMem * & 0.381 & 0.432 & 0.090 & 0.108 & 0.689 & 0.702 \\ 
DyGFormer \cite{yu2023towards} & \first{0.816 $\pm$ 0.005} & \first{0.798 $\pm$ 0.004} & 0.219 $\pm$ 0.017 & 0.224 $\pm$ 0.015 & 0.730 $\pm$ 0.002 & 0.752 $\pm$ 0.004 \\
TNCN \cite{zhang2024efficient} & \third{0.731 $\pm$ 0.001} & \third{0.718 $\pm$ 0.001} & \second{0.325 $\pm$ 0.003} & \second{0.377 $\pm$ 0.010} & \third{0.740 $\pm$ 0.002} & \third{0.762 $\pm$ 0.004} \\ 
TGN \cite{rossi2020temporal} & 0.435 $\pm$ 0.069 & 0.396 $\pm$ 0.060 & 0.313 $\pm$ 0.012 & 0.349 $\pm$ 0.020 & 0.607 $\pm$ 0.014 & 0.586 $\pm$ 0.037 \\
GraphMixer \cite{cong2023graphmixer} & 0.113 $\pm$ 0.003 & 0.118 $\pm$ 0.002 & \first{0.428 $\pm$ 0.019} & \first{0.521 $\pm$ 0.015} & \second{0.721 $\pm$ 0.005} & \second{0.763 $\pm$ 0.001} \\
\midrule 
\midrule
\textbf{Method} & \multicolumn{2}{c}{\textbf{tgbl-comment}} & \multicolumn{2}{c}{\textbf{tgbl-flight}} \\
 & Validation MRR & Test MRR & Validation MRR & Test MRR \\
\cmidrule(lr){1-5}
Base3 * & 0.426 & 0.450 & \third{0.809} & \third{0.794} \\ 
t-CoMem * & 0.341 & 0.447 & \first{0.846} & \first{0.840} \\ 
DyGFormer \cite{yu2023towards} & \third{0.613 $\pm$ 0.003} & \third{0.670 $\pm$ 0.001} & OOT & OOT \\ 
TNCN \cite{zhang2024efficient} & \second{0.642 $\pm$ 0.003} & \second{0.697 $\pm$ 0.006} & \second{0.831 $\pm$ 0.003} & \second{0.820 $\pm$ 0.004} \\ 
TGN \cite{rossi2020temporal} & 0.356 $\pm$ 0.019 & 0.379 $\pm$ 0.021 & 0.731 $\pm$ 0.01 & 0.705 $\pm$ 0.020 \\
GraphMixer \cite{cong2023graphmixer} & \first{0.701 $\pm$ 0.010} & \first{0.765 $\pm$ 0.009} & OOT & OOT \\
\cmidrule(lr){1-5}
\end{tabular}
\end{table*}

\subsection{Interpolation Models}

To combine the signals from EdgeBank, PopTrack, and t-CoMem, we define an interpolated score:

\begin{equation}
\text{score}_{\text{ Base3}}(u, v, t) = \alpha \cdot s_{\text{ EB}} + \beta \cdot s_{\text{ PT}} + \delta \cdot s_{\text{ CM}}
\end{equation}

where:
\begin{itemize}
    \item $s_{\text{ EB}} = s_{\text{ EB}}(u, v) =$ the EdgeBank score: 1 if $(u, v)$ has been observed before, 0 otherwise.
    \vskip .1in
    \item $s_{\text{ PT}} = s_{\text{ PT}}(v) =$ the PopTrack score for the destination node $v$: 1 if $v$ is in the top-$K$ nodes, 0 otherwise.
    \vskip .1in
    \item $s_{\text{ CM}} = s_{\text{ CM}}(u, v, t) =$ the t-CoMem score, which depends on the source's recent interactions and the decayed popularity of intermediate neighbors.
    \vskip .1in
    \item $(\alpha, \beta, \delta)$ are interpolation weights chosen based on the interpolation strategy.
\end{itemize}

\vskip .1in 

Base3 computes a weighted sum of EdgeBank, PopTrack, and t-CoMem's outputs according to an interpolation scheme. We experiment with three such schemes:

\vskip .1in 

\begin{itemize}
    \item \texttt{Uniform} assigns equal weight to each component ($\alpha = \beta = \delta = \frac{1}{3}$), assuming that EdgeBank, PopTrack, and t-CoMem are equally informative across all contexts.

    \vskip .1in

    \item \texttt{EB\_conf} weights the components based on EdgeBank confidence. This model is based on the assumption that if a potential edge that is being scored for prediction is already in the edge bank, its EdgeBank score is more significative than otherwise, and thus that repeated interactions are highly predictive when available. Considering the population process of the edge bank, this presents itself as a promising signal. If an edge exists in EdgeBank ($s_{\text{EB}}=1$), its contribution is thus up-weighted relative to the others. A more detailed insight into the weight repartition is made available in Appendix~\ref{app:interpo}.

    \vskip .1in

    \item \texttt{multi\_conf} extends \texttt{EB\_conf} by also factoring in the popularity of $v$ from PopTrack. When both EdgeBank and PopTrack show strong signals (i.e., the edge exists and the destination node is in the top-$K$ popular nodes), both their weights are increased. If neither signal is strong, t-CoMem is favored as a fallback. Likewise, more details on this model's weight formulation process is available in Appendix~\ref{app:interpo}.
\end{itemize}


\vskip .1in

We explore the performance of each model in our ablation studies, finding that confidence-based models achieve higher performance. Our proposed Base3 model uses \texttt{multi\_conf} by default, as it is the interpolation model which most strongly compounds the three components, as well as empirically performing the best.

\section{Experiments}

\subsection{Datasets}

We evaluate our proposed methods across the TGB benchmark, a collection of diverse benchmark datasets for robust and reproducible evaluation \cite{TGB}. TGB contains a range of 5 datasets for dynamic link prediction, which vary in size and surprise, guaranteeing a realistic overview of our models' performance across different settings. Here, \textit{surprise} refers to the metric developed by \cite{edgebank} which quantifies the degree of novelty in the test set of a temporal graph, relative to the training set. It is proportional to the difficulty of predicting dynamic links on the dataset, and is defined as follows: 

\begin{equation}
\textit{surprise} = \frac{\left| E_{\text{test}} \setminus E_{\text{train}} \right|}{\left| E_{\text{test}} \right|}
\end{equation}  

\vskip .1in 

 A breakdown of the TGB datasets for dynamic link prediction, and their level of surprise, is put forth in Table~\ref{tab:tgb}. 

\begin{table}[H]
\centering
\caption{Overview of TGB datasets \cite{TGB}}
\label{tab:tgb}
\begin{tabular}{l c c c c}
\toprule
\textbf{Name} & \textbf{\#Nodes} & \textbf{\#Edges} & \textbf{\#Steps} & \textbf{Surprise} \\
\midrule
\texttt{wiki-v2}    & 9,227     & 157,474     & 152,757    & 0.108 \\
\texttt{review-v2}  & 352,637   & 4,873,540   & 6,865      & 0.987 \\
\texttt{coin}       & 638,486   & 22,809,486  & 1,295,720  & 0.120 \\
\texttt{comment}    & 994,790   & 44,314,507  & 30,998,030 & 0.823 \\
\texttt{flight}     & 18,143    & 67,169,570  & 1,385      & 0.024 \\
\bottomrule
\end{tabular}
\end{table}

\subsection{Results}

\begin{figure}[ht]
    \centering
    \includegraphics[width=1.2\linewidth]{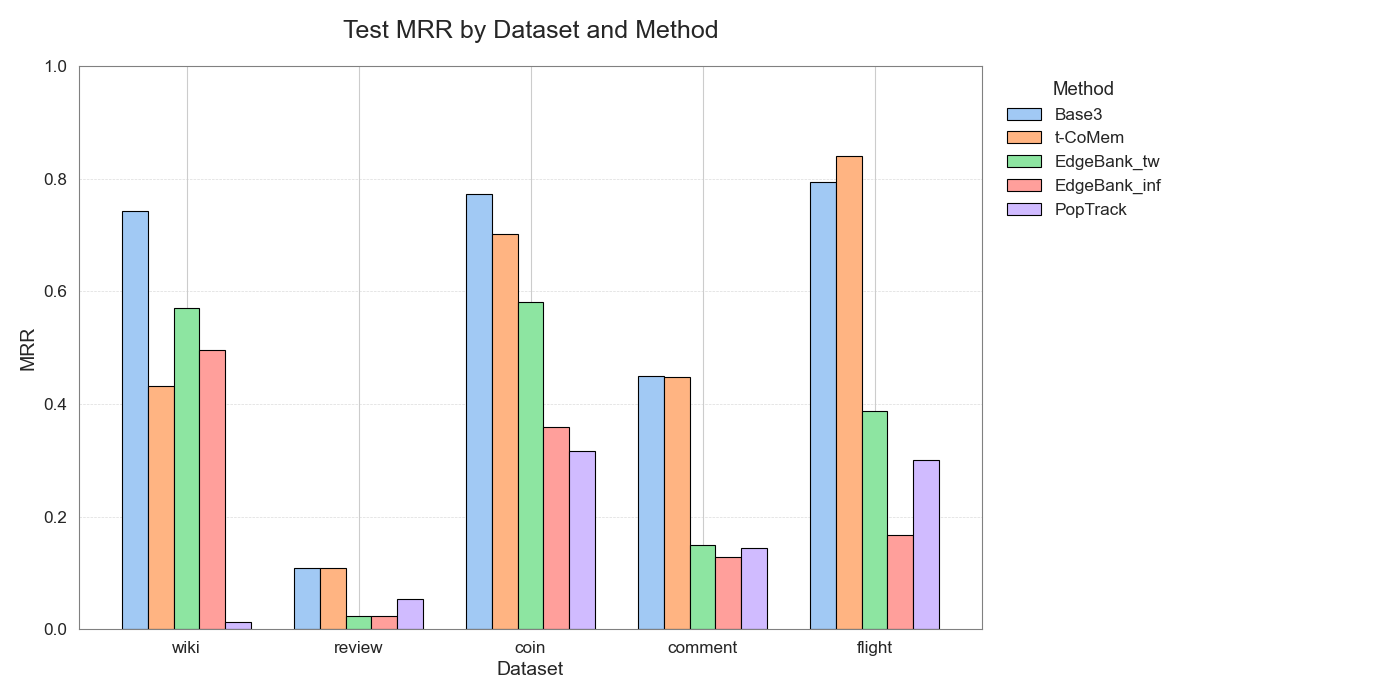} 
    \caption{Test MRR by dataset for non-learnable dynamic link prediction methods (Base3 model components).}
    \label{fig:test_mrr}
\end{figure}

We report performance using the standard evaluation metric in TGB, Mean Reciprocal Rank (MRR) on both the validation and test sets for our main experiments -- we also look at Area Under the Receiver Operating Characteristic curve (AUROC) scores, in our ablation studies. For each dataset, we compare our proposed models, \textbf{t-CoMem} and \textbf{Base3}, against existing non-learnable baselines (EdgeBank and PopTrack \cite{edgebank, PopTrack}) as well as state-of-the-art trainable models DyGFormer, TNCN, TGN, and GraphMixer \cite{yu2023towards, rossi2020temporal, cong2023graphmixer, zhang2024efficient}. A key point to note about these results is that, as with original reports of EdgeBank’s performance, we do not report standard deviations. This is because Base3 and its components are entirely deterministic: their predictions do not vary across runs or random seeds. Consequently, repeated executions yield identical outputs, and we therefore present only single-run results.

\vskip .1in

Figure~\ref{fig:test_mrr} presents the MRR scores across the five datasets of the non-learnable baselines. Comparing Base3's individual components against each other strongly highlights t-CoMem as taking the lead in performance, while also putting forward the strength of compounding them together. Indeed, Base3 always outperforms its individual components by a considerable margin, except on \texttt{tgbl-flight}, on which t-CoMem itself performs best. Thus, our mixture-of-experts approach consistently outperforms existing baselines while maintaining strong levels of comprehensibility. While a non-linear learning scheme may further improve this and push Base3 to consistently outperform t-CoMem, such a method would limit interpretability. Notably, our results show: 
\vskip .1in

\begin{itemize}
    \item On \texttt{tgbl-wiki-v2}, Base3 achieves a high test MRR (0.743). Both EdgeBank and PopTrack perform more poorly on this dataset, which exhibits modest novelty (surprise = 0.108). Base3’s success here stems from its ability to yield EdgeBank's recurrent pattern recognition strengths, while falling back on t-CoMem’s neighborhood-aware scores when explicit memorization (EdgeBank) or popularity (PopTrack) fail. 
    
    \vskip .1in 

    \item With the highest surprise score (0.987), the \texttt{tgbl-review-v2} dataset tests inductive generalization to unseen interactions. Base3 significantly improves over both EdgeBank and PopTrack, reaching 0.108 MRR. The improvement is largely due to t-CoMem, which compensates for EdgeBank’s total failure in inductive settings and for PopTrack’s limited scope. Here, Base3 benefits from its adaptive weighting strategy (\texttt{multi\_conf}).

    \vskip .1in 

    \item On \texttt{tgbl-coin}, Base3 outperforms its individual components, and is closely follows by t-CoMem, while PopTrack and EdgeBank are performing considerably (more than 10\%) lower. \texttt{Tgbl-coin} has a considerably low surprise rate as well (0.120), coherent with other datasets on which Base3 excels.  
    
    \vskip .1in 

    \item \texttt{Tgbl-comment} is another high-suprise dataset (0.823), on which Base3 outperforms the two baselines by a significant margin. Notably, t-CoMem alone nearly matches Base3, suggesting that memory of co-occurrence patterns is critical in this setting. EdgeBank and PopTrack underperform due to limited recurrence and fluctuating node popularity. This supports t-CoMem's importance and standalone strength. 

    \vskip .1in 

    \item \texttt{Tgbl-flight} is the only dataset where Base3 does not outperform its best component. t-CoMem alone achieves the highest MRR among all non-learnable methods (with a score of 84\%). Base3 also performs strongly, despite both EdgeBank and PopTrack's MRRs staying below 40\%. We conjecture that the extreme node and edge volume, along with a very low surprise score (0.024), make recurrence highly informative, and t-CoMem’s source-aware memory is a strong signal that gets diluted by PopTrack and EdgeBank. 
\end{itemize}

\begin{table*}[ht]
\centering
\small
\caption{Ablation studies of Base3 on \texttt{tgbl-wiki-v2} under varying co-occurence weights, memory spans and K values. The best performance is boldened and yields the default parameter combination used in other experiments.}
\label{wiki_abl}
\begin{tabular}{lllrr}
\toprule
\textbf{Memory span} & \textbf{Co-occurrence weight} & \textbf{K} & \textbf{MRR$_\text{val}$} & \textbf{MRR$_\text{test}$} \\
\midrule 

0.01 & 0.25 & 100 & 0.212 & 0.214 \\ 
  0.01 & 0.5 & 100 & 0.233 & 0.222 \\
  0.01 & 0.75 & 100 & 0.233 & 0.222 \\ 
  0.01 & 1.0 & 100 & 0.233 & 0.222  \\ 
  \midrule 
  0.1 & 0.25 & 100 & 0.215 & 0.218  \\ 
  0.1 & 0.5 & 100 & 0.243 & 0.228  \\ 
  0.1 & 0.75 & 100 & 0.243 & 0.228  \\ 
  0.1 & 1.0 & 100 & 0.243 & 0.228  \\ 
  \midrule 
  1.0 & 0.25 & 100 & 0.214 & 0.218  \\ 
  1.0 & 0.5 & 100 & 0.244 & 0.229  \\ 
  1.0 & 0.75 & 100 & 0.245 & 0.229 \\ 
  1.0 & 1.0 & 100 & 0.245 & 0.229 \\
  \midrule 
  \midrule 
  0.01 & 0.25 & 1000 & 0.399 & 0.446 \\ 
  0.01 & 0.5 & 1000 & 0.644 & 0.639 \\
  0.01 & 0.75 & 1000 & 0.649 & 0.644 \\ 
  0.01 & 1.0 & 1000 & 0.649 & 0.644  \\ 
  \midrule 
  0.1 & 0.25 & 1000 & 0.392 & 0.453  \\ 
  0.1 & 0.5 & 1000 & 0.721 & 0.737  \\ 
  0.1 & 0.75 & 1000 & 0.727 & 0.743  \\ 
  \textbf{0.1} & \textbf{1.0} & \textbf{1000} & \textbf{0.727} & \textbf{0.743}  \\ 
  \midrule 
  1.0 & 0.25 & 1000 & 0.387 & 0.436  \\ 
  1.0 & 0.5 & 1000 & 0.729 & 0.720  \\ 
  1.0 & 0.75 & 1000 & 0.736 & 0.727 \\ 
  1.0 & 1.0 & 1000 & 0.736 & 0.727 \\
\bottomrule
\end{tabular}
\end{table*}

\vskip .1in

Having established Base3 outperforms non-trained baselines, we look to determine whether it can compete with complex learnable models as well. Specifically, in Table~\ref{fullMRR}, we compare our ensemble model \textbf{Base3} with the state-of-the-art DyGFormer, TNCN, TGN, and GraphMixer \cite{yu2023towards, rossi2020temporal, cong2023graphmixer, zhang2024efficient}, models which currently stand at the lead of the TGB leaderboards for dynamic link predictions. Table~\ref{fullMRR} shows that Base3 achieves consistently strong performance, outperforming diverse models across multiple settings. Notably:

\begin{itemize}
    \item On \texttt{tgbl-wiki-v2}, Base3 ranks second overall, trailing only DyGFormer. While EdgeBank and PopTrack each perform modestly in isolation, their combination with t-CoMem in Base3 captures both recurring and contextual patterns more effectively. Remarkably, it outperforms TGN, TNCN, and GraphMixer, despite being completely training-free. This suggest complex graph analysis may be unnecessary in contexts where simple pattern recognition already excels (especially considering \texttt{wiki}'s low surprise). 
    
    \vskip .1in 
    
    \item On \texttt{tgbl-review-v2} and \texttt{tgbl-comment}, the highest-surprise sets, neural models beat Base3. Base3 particularly underperforms on \texttt{tgbl-review-v2}, suggesting that in high-surprise contexts, simple pattern recognition may not be enough for robust link prediction.

    \vskip .1in

    \item On \texttt{tgbl-coin}, Base3 delivers the highest test MRR across all methods, including state-of-the-art deep architectures. The dataset's moderate surprise score and consistent structure favor methods that blend memorization (EdgeBank) with temporal context (t-CoMem). Base3’s design capitalizes on this by assigning meaningful weight to recurrence and co-occurrence without being misled by volatile popularity spikes.

    \vskip .1in

    \item On \texttt{tgbl-flight}, not only is Base3 in the top 3 ranking, but t-CoMem itself ranks first, outperforming the TNCN and TGN models. This dataset, with its vast scale and low surprise score (0.024), is highly structured—historical recurrence is strongly predictive. In such cases, t-CoMem’s ability to retain fine-grained, source-aware memory becomes dominant. Given the large size of \texttt{tgbl-flight}, however, more consuming models (DyGFormer and GraphMixer) were unable to run to completion due to computational limitations, further emphasizing the efficiency advantage of our approach. The high performance on the behalf of both t-CoMem and Base3 is especially interesting considering how low each of the baselines (EdgeBank and PopTrack) scores, comparatively. 
\end{itemize}

These results confirm that combining recurrence (EdgeBank), popularity (PopTrack), and source-aware co-occurrence (t-CoMem) yields a more generalizable predictor than relying on any single heuristic alone.

\begin{table*}[ht]
\centering
\small
\caption{Ablation studies of Base3 on \texttt{tgbl-wiki-v2} and \texttt{tgbl-review-v2} under varying interpolation models}
\label{interpo_abl}
\begin{tabular}{lrrrrcccc}
\toprule
\textbf{Model} & \textbf{Memory span} & \textbf{Co-occurrence weight} & \textbf{K} 
& \multicolumn{2}{c}{\textbf{tgbl-wiki-v2}} 
& \multicolumn{2}{c}{\textbf{tgbl-review-v2}} \\
\cmidrule(r){5-6} \cmidrule(l){7-8}
& & & & \textbf{MRR$_\text{val}$} & \textbf{MRR$_\text{test}$} & \textbf{MRR$_\text{val}$} & \textbf{MRR$_\text{test}$} \\
\midrule
\multirow{2}{*}{Uniform}
  & 0.01 & 1.0 & 1000 & 0.649 & 0.644 & 0.053 & 0.083\\ 
  & \textbf{0.1} & \textbf{1.0} & \textbf{1000} & \textbf{0.727} & \textbf{0.743} & \textbf{0.048} & \textbf{0.084} \\ 
  & 1.0 & 1.0 & 1000 & 0.736 & 0.727 & 0.034 & 0.047 \\ 
\midrule 
\multirow{3}{*}{Multi\_conf}
  & 0.01 & 1.0 & 1000 & 0.649 & 0.644 & 0.102 & 0.108 \\ 
  & \textbf{0.1} & \textbf{1.0} & \textbf{1000} & \textbf{0.727} & \textbf{0.743} & \textbf{0.101} & \textbf{0.108} \\ 
  & 1.0 & 1.0 & 1000 & 0.736 & 0.727 & 0.101 & 0.107 \\
\midrule
\multirow{3}{*}{EB\_conf}
  & 0.01 & 1.0 & 1000 & 0.719 & 0.686 & 0.053 & 0.084 \\ 
  & \textbf{0.1} & \textbf{1.0} & \textbf{1000} & \textbf{0.749} & \textbf{0.752} & \textbf{0.048} & \textbf{0.084} \\ 
  & 1.0 & 1.0 & 1000 & 0.742 & 0.738 & 0.035 & 0.047 \\ 
\bottomrule
\end{tabular}
\end{table*}

\subsection{Ablation Studies}

To understand the role of each of Base3's hyper-parameters, we perform some ablation studies, as illustrated in Table~\ref{wiki_abl}. These experiments are ran with the \texttt{multi\_conf} interpolation scheme. There are three hyperparameters, each of which has a considerable effect on Base3's performance:

\begin{itemize}
    \item The first hyperparameter is the memory span. This is equivalent to the same hyperparameter introduced for EdgeBank \cite{edgebank}, and determines how far back the memory reaches, as a percentage of the entire history. We explore with memory span values $[0.01, 0.1, 1.0]$ and find that, considerably so, a higher memory span increases performance. This is generally true, though we see a slight decrease from a 0.1 span to a 1.0 one, when K is larger. As such, we find the optimal memory span to be either 0.1 or 1.0, fixing it to 0.1 as that is optimal under optimal choices for other hyperparameters.
    
    \vskip .1in

    \item The second hyperparameter is co-occurence weight, and stems from t-CoMem's logic of weighting co-occurence scores. Similarly, by trying different weights in \\ $[0.25, 0.50, 0.75, 1.0]$, we find that a higher co-occurence weight yields higher MRR scores, and thus fix Base3's default to 1.0. 

    \vskip .1in 

    \item The last hyperparameter is the K value, which stems from PopTrack's logic. In the PopTrack model, a score is positive if the destination is in the top-$K$ most popular nodes. Interestingly, while the original authors of the model reported that the optimal $K$-value was 100 \cite{PopTrack}, we find that Base3's performance with $K=100$ is much lower (about 50\% so) than with $K=1000$. We suppose this stems from the different ways in which Base3 leverages these top-$K$ nodes and PopTrack scores themselves. 

    \vskip .1in 

    \item As such, we select the optimal combination of hyperparameters and set Base3's defaults to them -- specifically, we set a memory span of $0.1$, a co-occurence weight of $1.0$ and a $K$-value of 1000. 
\end{itemize}

\begin{table*}[ht]
\centering
\small
\caption{AUROC performance of each model under varying negative sampling strategies on \texttt{tgbl-wiki-v2}.}
\label{NS_abl}
\begin{tabular}{llcc}
\toprule
\textbf{Negative Sampling Strategy} & \textbf{Model} & \textbf{AUROC$_\text{val}$} & \textbf{AUROC$_\text{test}$} \\
\midrule
\multirow{4}{*}{Random} 
    & \textbf{Base3} & \textbf{0.922} & \textbf{0.915} \\
    & t-CoMem & 0.914 & 0.909 \\
    & EdgeBank$_\text{tw}$ & 0.875 & 0.866 \\
    & PopTrack & 0.551 & 0.560 \\
\midrule
\multirow{4}{*}{Inductive} 
    & Base3 & 0.808 & 0.721 \\
    & \textbf{t-CoMem} & \textbf{0.923} & \textbf{0.800} \\
    & EdgeBank$_\text{tw}$ & 0.876 & 0.421 \\
    & PopTrack & 0.567 & 0.498 \\
\midrule
\multirow{4}{*}{Historical} 
    & \textbf{Base3} & \textbf{0.797} & \textbf{0.781} \\
    & t-CoMem & 0.768 & 0.750 \\
    & EdgeBank$_\text{tw}$ & 0.740 & 0.775 \\
    & PopTrack & 0.505 & 0.488 \\
\bottomrule
\end{tabular}
\end{table*}

\subsection{Comparing Interpolation Models}

We then look at our three proposed interpolation models in more detail. Table~\ref{interpo_abl} illustrates our results, looking at the \texttt{uniform}, \texttt{multi\_conf} and \texttt{EB\_conf} models on both the \texttt{tgbl-wiki-v2} and \texttt{tgbl-review} datasets. As can be observed, \texttt{uniform} and \texttt{multi\_conf} yield the same performance on \texttt{tgbl-wiki-v2}, both being lower than \texttt{EB\_conf}. This trend, however, is not reproduced in other datasets, as we empirically observed, and report for \texttt{tgbl-review}. Indeed, on \texttt{tgbl-review}, both \texttt{uniform} and \texttt{EB\_conf} perform a few points lower than \texttt{multi\_conf}. Considering \texttt{tgbl-review} is harder than \texttt{tgbl-wiki-v2} (it has a higher surprise), these insights are quite important, and determined us setting Base3's default strategy to \texttt{multi\_conf} to produce a more robust model.

\subsection{Studies under different negative sampling strategies}

As highlighted previously, the choice of the negative sampling strategy used is of great importance when evaluating a temporal graph model. As such, we look at Base3's performance under the three different negative sampling strategies principally used in the literature--random sampling, as well as inductive and historical. 

\vskip .1in

As put forth in our ablation studies, the settings used in these experiments are set to the optimal empirical combination: the memory span is set to 0.1, the co-occurence weight to 1.0, and the $K$ value to 1000. The interpolation model used is \texttt{multi\_conf}. Here, we look at the model's Area Under the Receiver Operating Characteristic (AUROC) performance, as this is the metric used in the literature that developed inductive and historical negative sampling, namely \cite{edgebank}.

\vskip .1in

As shown in Table~\ref{NS_abl}, the existing baselines exhibit a marked drop in test MRR when the negative sampling strategy is altered—most notably under inductive sampling. EdgeBank${_\text{ tw}}$ performs particularly poorly, achieving less than half its random sampling performance, while PopTrack suffers a nearly 10\% decrease. In stark contrast, both t-CoMem and Base3 maintain robust performance across \textbf{all} sampling settings, with AUROC scores consistently above 72\%. This resilience under inductive sampling highlights the effectiveness of their design and confirms the success of our core objective: endowing EdgeBank$_{\text{ tw}}$ with inductive generalization through the integration of memory-based co-occurrence (t-CoMem) and popularity-aware interpolation (Base3).

\section{Conclusion}
We introduce a lightweight yet effective framework for enhancing non-learnable temporal link prediction models with inductive capabilities. By integrating co-occurrence-aware memory (t-CoMem) and popularity-driven reasoning into the EdgeBank baseline, we developed Base3, a training-free ensemble that combines recurrence, global popularity, and local temporal context into a unified scoring mechanism. Extensive evaluation across diverse datasets and under challenging negative sampling regimes demonstrates that Base3 not only outperforms traditional baselines but also rivals the performance of state-of-the-art deep learning models—without requiring training, tuning, or backpropagation. Notably, its strong performance under inductive and historical sampling confirms the success of our central objective: to enable robust generalization to unseen nodes and interactions. These results advocate for a reevaluation of complexity in temporal graph learning, suggesting that well-designed non-parametric models can offer a scalable, interpretable, and competitive alternative for dynamic link prediction in real-world applications.

\section{Acknowledgements}
This research was enabled in part by compute resources provided by Mila (mila.quebec). 

\bibliographystyle{ACM-Reference-Format}
\bibliography{references}


\begin{thebibliography}{23}


\ifx \showCODEN    \undefined \def \showCODEN     #1{\unskip}     \fi
\ifx \showDOI      \undefined \def \showDOI       #1{#1}\fi
\ifx \showISBNx    \undefined \def \showISBNx     #1{\unskip}     \fi
\ifx \showISBNxiii \undefined \def \showISBNxiii  #1{\unskip}     \fi
\ifx \showISSN     \undefined \def \showISSN      #1{\unskip}     \fi
\ifx \showLCCN     \undefined \def \showLCCN      #1{\unskip}     \fi
\ifx \shownote     \undefined \def \shownote      #1{#1}          \fi
\ifx \showarticletitle \undefined \def \showarticletitle #1{#1}   \fi
\ifx \showURL      \undefined \def \showURL       {\relax}        \fi
\providecommand\bibfield[2]{#2}
\providecommand\bibinfo[2]{#2}
\providecommand\natexlab[1]{#1}
\providecommand\showeprint[2][]{arXiv:#2}

\bibitem[Barab{\'a}si and Albert(1999)]%
        {barabasi1999emergence}
\bibfield{author}{\bibinfo{person}{Albert-L{\'a}szl{\'o} Barab{\'a}si} {and} \bibinfo{person}{R{\'e}ka Albert}.} \bibinfo{year}{1999}\natexlab{}.
\newblock \showarticletitle{Emergence of scaling in random networks}.
\newblock \bibinfo{journal}{\emph{Science}} \bibinfo{volume}{286}, \bibinfo{number}{5439} (\bibinfo{year}{1999}), \bibinfo{pages}{509--512}.
\newblock


\bibitem[Cong et~al\mbox{.}(2023)]%
        {cong2023graphmixer}
\bibfield{author}{\bibinfo{person}{Weilin Cong}, \bibinfo{person}{Si Zhang}, \bibinfo{person}{Jian Kang}, \bibinfo{person}{Baichuan Yuan}, \bibinfo{person}{Hao Wu}, \bibinfo{person}{Xin Zhou}, \bibinfo{person}{Hanghang Tong}, {and} \bibinfo{person}{Mehrdad Mahdavi}.} \bibinfo{year}{2023}\natexlab{}.
\newblock \showarticletitle{Do we really need complicated model architectures for temporal networks?}. In \bibinfo{booktitle}{\emph{International Conference on Learning Representations (ICLR)}}.
\newblock


\bibitem[Cornell et~al\mbox{.}(2025)]%
        {cornell2025power}
\bibfield{author}{\bibinfo{person}{Filip Cornell}, \bibinfo{person}{Oleg Smirnov}, \bibinfo{person}{Gabriela~Zarzar Gandler}, {and} \bibinfo{person}{Lele Cao}.} \bibinfo{year}{2025}\natexlab{}.
\newblock \bibinfo{title}{On the Power of Heuristics in Temporal Graphs}.
\newblock \bibinfo{howpublished}{arXiv preprint}.
\newblock


\bibitem[Daniluk and Dabrowski(2024)]%
        {PopTrack}
\bibfield{author}{\bibinfo{person}{Michal Daniluk} {and} \bibinfo{person}{Jacek Dabrowski}.} \bibinfo{year}{2024}\natexlab{}.
\newblock \bibinfo{title}{Temporal graph models fail to capture global temporal dynamics}.
\newblock \bibinfo{howpublished}{arXiv preprint, ICLR 2024 (withdrawn submission)}.
\newblock
\showeprint[openreview]{9kLDrE5rsW}
\newblock
\shownote{Available at \url{https://openreview.net/forum?id=9kLDrE5rsW}}.


\bibitem[Dileo and Zignani(2024)]%
        {dileo2024link}
\bibfield{author}{\bibinfo{person}{Manuel Dileo} {and} \bibinfo{person}{Matteo Zignani}.} \bibinfo{year}{2024}\natexlab{}.
\newblock \showarticletitle{Link prediction heuristics for temporal graph benchmark}. In \bibinfo{booktitle}{\emph{Proceedings of the European Symposium on Artificial Neural Networks, Computational Intelligence and Machine Learning (ESANN 2024)}}.
\newblock
\urldef\tempurl%
\url{https://www.esann.org/sites/default/files/proceedings/2024/ES2024-141.pdf}
\showURL{%
\tempurl}


\bibitem[Easley and Kleinberg(2010)]%
        {easley2010networks}
\bibfield{author}{\bibinfo{person}{David Easley} {and} \bibinfo{person}{Jon Kleinberg}.} \bibinfo{year}{2010}\natexlab{}.
\newblock \bibinfo{booktitle}{\emph{Networks, Crowds, and Markets: Reasoning About a Highly Connected World}}.
\newblock \bibinfo{publisher}{Cambridge University Press}.
\newblock


\bibitem[Hamilton et~al\mbox{.}(2017)]%
        {hamilton2017}
\bibfield{author}{\bibinfo{person}{William~L. Hamilton}, \bibinfo{person}{Rex Ying}, {and} \bibinfo{person}{Jure Leskovec}.} \bibinfo{year}{2017}\natexlab{}.
\newblock \showarticletitle{Inductive Representation Learning on Large Graphs}. In \bibinfo{booktitle}{\emph{Advances in Neural Information Processing Systems (NeurIPS)}}, Vol.~\bibinfo{volume}{30}.
\newblock
\urldef\tempurl%
\url{https://papers.nips.cc/paper_files/paper/2017/hash/5dd9db5e033da9c6fb5ba83c7a7ebea9-Abstract.html}
\showURL{%
\tempurl}


\bibitem[Huang et~al\mbox{.}(2023)]%
        {TGB}
\bibfield{author}{\bibinfo{person}{Shenyang Huang}, \bibinfo{person}{Farimah Poursafaei}, \bibinfo{person}{Jacob Danovitch}, \bibinfo{person}{Matthias Fey}, \bibinfo{person}{Weihua Hu}, \bibinfo{person}{Emanuele Rossi}, \bibinfo{person}{Jure Leskovec}, \bibinfo{person}{Michael Bronstein}, \bibinfo{person}{Guillaume Rabusseau}, {and} \bibinfo{person}{Reihaneh Rabbany}.} \bibinfo{year}{2023}\natexlab{}.
\newblock \showarticletitle{Temporal Graph Benchmark for Machine Learning on Temporal Graphs}.
\newblock \bibinfo{journal}{\emph{Advances in Neural Information Processing Systems}} (\bibinfo{year}{2023}).
\newblock


\bibitem[Huang et~al\mbox{.}(2024)]%
        {huang2024utg}
\bibfield{author}{\bibinfo{person}{Shenyang Huang}, \bibinfo{person}{Farimah Poursafaei}, \bibinfo{person}{Reihaneh Rabbany}, \bibinfo{person}{Guillaume Rabusseau}, {and} \bibinfo{person}{Emanuele Rossi}.} \bibinfo{year}{2024}\natexlab{}.
\newblock \bibinfo{title}{UTG: Towards a Unified View of Snapshot and Event Based Models for Temporal Graphs}.
\newblock \bibinfo{howpublished}{arXiv preprint}.
\newblock
\urldef\tempurl%
\url{https://arxiv.org/abs/2407.12269}
\showURL{%
\tempurl}
\newblock
\shownote{arXiv:2407.12269 [cs.LG]}.


\bibitem[Kazemi et~al\mbox{.}(2020)]%
        {kazemi2020representation}
\bibfield{author}{\bibinfo{person}{Seyed~Mehran Kazemi}, \bibinfo{person}{Rishab Goel}, \bibinfo{person}{Shikhar Jain}, \bibinfo{person}{Ivan Kobyzev}, \bibinfo{person}{Luke Sethi}, \bibinfo{person}{Peter Forsyth}, {and} \bibinfo{person}{Pascal Poupart}.} \bibinfo{year}{2020}\natexlab{}.
\newblock \showarticletitle{Representation learning for dynamic graphs: A survey}.
\newblock \bibinfo{journal}{\emph{Journal of Machine Learning Research}} \bibinfo{volume}{21}, \bibinfo{number}{70} (\bibinfo{year}{2020}), \bibinfo{pages}{1--73}.
\newblock


\bibitem[Kipf and Welling(2017)]%
        {kipf2017}
\bibfield{author}{\bibinfo{person}{Thomas~N Kipf} {and} \bibinfo{person}{Max Welling}.} \bibinfo{year}{2017}\natexlab{}.
\newblock \showarticletitle{Semi-Supervised Classification with Graph Convolutional Networks}. In \bibinfo{booktitle}{\emph{International Conference on Learning Representations (ICLR)}}.
\newblock
\urldef\tempurl%
\url{https://arxiv.org/abs/1609.02907}
\showURL{%
\tempurl}


\bibitem[Liben-Nowell and Kleinberg(2007)]%
        {liben2007link}
\bibfield{author}{\bibinfo{person}{David Liben-Nowell} {and} \bibinfo{person}{Jon Kleinberg}.} \bibinfo{year}{2007}\natexlab{}.
\newblock \showarticletitle{The link-prediction problem for social networks}. In \bibinfo{booktitle}{\emph{Journal of the American society for information science and technology}}, Vol.~\bibinfo{volume}{58}. \bibinfo{publisher}{Wiley Online Library}, \bibinfo{pages}{1019--1031}.
\newblock


\bibitem[Lichtenwalter et~al\mbox{.}(2010)]%
        {lichtenwalter2010new}
\bibfield{author}{\bibinfo{person}{Ryan~N Lichtenwalter}, \bibinfo{person}{Jake~T Lussier}, {and} \bibinfo{person}{Nitesh~V Chawla}.} \bibinfo{year}{2010}\natexlab{}.
\newblock \showarticletitle{New perspectives and methods in link prediction}.
\newblock \bibinfo{journal}{\emph{Proceedings of the 16th ACM SIGKDD international conference on Knowledge discovery and data mining}} (\bibinfo{year}{2010}), \bibinfo{pages}{243--252}.
\newblock


\bibitem[Mohammadzadeh(2024)]%
        {mohammadzadeh2024temporal}
\bibfield{author}{\bibinfo{person}{Shahrad Mohammadzadeh}.} \bibinfo{year}{2024}\natexlab{}.
\newblock \bibinfo{title}{Temporal Collaborative Filtering: Enhancing EdgeBank with Inductive Capabilities}.  (\bibinfo{year}{2024}).
\newblock
\newblock
\shownote{Manuscript, not published}.


\bibitem[Pareja(2020)]%
        {pareja2020evolvegcn}
\bibfield{author}{\bibinfo{person}{Amber et~al. Pareja}.} \bibinfo{year}{2020}\natexlab{}.
\newblock \showarticletitle{EvolveGCN: Evolving Graph Convolutional Networks for Dynamic Graphs}. In \bibinfo{booktitle}{\emph{Proceedings of the AAAI Conference on Artificial Intelligence}}, Vol.~\bibinfo{volume}{34}. \bibinfo{pages}{5363--5370}.
\newblock


\bibitem[Poursafaei et~al\mbox{.}(2022)]%
        {edgebank}
\bibfield{author}{\bibinfo{person}{Farimah Poursafaei}, \bibinfo{person}{Shenyang Huang}, \bibinfo{person}{Kellin Pelrine}, {and} \bibinfo{person}{Reihaneh Rabbany}.} \bibinfo{year}{2022}\natexlab{}.
\newblock \showarticletitle{Towards Better Evaluation for Dynamic Link Prediction}.
\newblock \bibinfo{journal}{\emph{arXiv preprint arXiv:2207.10128}} (\bibinfo{year}{2022}).
\newblock
\urldef\tempurl%
\url{https://arxiv.org/abs/2207.10128}
\showURL{%
\tempurl}


\bibitem[Rossi et~al\mbox{.}(2020)]%
        {rossi2020temporal}
\bibfield{author}{\bibinfo{person}{Emanuele Rossi}, \bibinfo{person}{Ben Chambers}, \bibinfo{person}{Rex Ying}, \bibinfo{person}{Michael Bronstein}, {and} \bibinfo{person}{Bruno Ribeiro}.} \bibinfo{year}{2020}\natexlab{}.
\newblock \showarticletitle{Temporal Graph Networks for Deep Learning on Dynamic Graphs}.
\newblock \bibinfo{journal}{\emph{arXiv preprint arXiv:2006.10637}} (\bibinfo{year}{2020}).
\newblock


\bibitem[Skarding et~al\mbox{.}(2021)]%
        {skarding2021foundations}
\bibfield{author}{\bibinfo{person}{Jo Skarding}, \bibinfo{person}{Bogdan Gabrys}, {and} \bibinfo{person}{Katarzyna Musial}.} \bibinfo{year}{2021}\natexlab{}.
\newblock \showarticletitle{Foundations and modelling of dynamic graphs}. In \bibinfo{booktitle}{\emph{Proceedings of the 2021 International Joint Conference on Neural Networks (IJCNN)}}. IEEE, \bibinfo{pages}{1--8}.
\newblock


\bibitem[Trivedi et~al\mbox{.}(2019)]%
        {trivedi2019}
\bibfield{author}{\bibinfo{person}{Rakshit Trivedi}, \bibinfo{person}{Mehrdad Farajtabar}, \bibinfo{person}{Parnam Biswal}, {and} \bibinfo{person}{Hongyuan Zha}.} \bibinfo{year}{2019}\natexlab{}.
\newblock \showarticletitle{DyRep: Learning Representations over Dynamic Graphs}. In \bibinfo{booktitle}{\emph{International Conference on Learning Representations (ICLR)}}.
\newblock
\urldef\tempurl%
\url{https://openreview.net/forum?id=HylMyhR5tm}
\showURL{%
\tempurl}


\bibitem[Veličković et~al\mbox{.}(2018)]%
        {velickovic2018}
\bibfield{author}{\bibinfo{person}{Petar Veličković}, \bibinfo{person}{Guillem Cucurull}, \bibinfo{person}{Arantxa Casanova}, \bibinfo{person}{Adriana Romero}, \bibinfo{person}{Pietro Lio}, {and} \bibinfo{person}{Yoshua Bengio}.} \bibinfo{year}{2018}\natexlab{}.
\newblock \showarticletitle{Graph Attention Networks}. In \bibinfo{booktitle}{\emph{International Conference on Learning Representations (ICLR)}}.
\newblock


\bibitem[Xu et~al\mbox{.}(2020)]%
        {xu2020inductive}
\bibfield{author}{\bibinfo{person}{Da Xu}, \bibinfo{person}{Chuanwei Ruan}, \bibinfo{person}{Evren Korpeoglu}, \bibinfo{person}{Sushant Kumar}, {and} \bibinfo{person}{Kannan Achan}.} \bibinfo{year}{2020}\natexlab{}.
\newblock \showarticletitle{Inductive Representation Learning on Temporal Graphs}. In \bibinfo{booktitle}{\emph{International Conference on Learning Representations (ICLR)}}.
\newblock


\bibitem[Yu et~al\mbox{.}(2023)]%
        {yu2023towards}
\bibfield{author}{\bibinfo{person}{Le Yu}, \bibinfo{person}{Leilei Sun}, \bibinfo{person}{Bowen Du}, {and} \bibinfo{person}{Weifeng Lv}.} \bibinfo{year}{2023}\natexlab{}.
\newblock \showarticletitle{Towards Better Dynamic Graph Learning: New Architecture and Unified Library}. In \bibinfo{booktitle}{\emph{Advances in Neural Information Processing Systems}}, Vol.~\bibinfo{volume}{36}. \bibinfo{pages}{67686--67700}.
\newblock


\bibitem[Zhang et~al\mbox{.}(2024)]%
        {zhang2024efficient}
\bibfield{author}{\bibinfo{person}{Xiaohui Zhang}, \bibinfo{person}{Yanbo Wang}, \bibinfo{person}{Xiyuan Wang}, {and} \bibinfo{person}{Muhan Zhang}.} \bibinfo{year}{2024}\natexlab{}.
\newblock \bibinfo{title}{Efficient Neural Common Neighbor for Temporal Graph Link Prediction}.
\newblock
\newblock
\showeprint[arxiv]{2406.07926}~[cs.LG]


\end{thebibliography}

\newpage
\appendix

\section{Interpolation Models}
\label{app:interpo}

Here, we provide a more detailed overview of the \texttt{EB\_conf} and \texttt{multi\_conf} interpolation models, specifically regarding the weighting mechanism. These schemes assign different value to the weight vector $[\alpha, \beta, \delta]$ where $\alpha$ is the weight given to the EdgeBank score, $\beta$ that to PopTrack, and $\delta$ that to t-CoMem.

\vskip .1in 

The initial weighting scheme is \texttt{uniform}, which simply linearly interpolates between the three scores by giving each of them a weight of $\frac13$. It is the most naive and uniformly weight-assigning approach. 

\vskip .1in 

\texttt{EB\_conf(eb\_score)} is a confidence-based weighting scheme that is centered around EdgeBank. It uses one confidence signal, \texttt{eb\_score} $\in [0,1]$, which is the discrete score given by the EdgeBank module; 1 if the edge in question is present in the edgebank, and 0 otherwise. Given this signal, the scheme choose between two weight vectors: $w_{\text{ conf}} = [0.5,0.2,0.3]$ and $w_{\text{ not}} = [0.2, 0.3, 0.5]$. Essentially, the logic of these two vectors is as follows: if EdgeBank is confident, it should be the principal component relied on. We simply equate \textit{principal} to $\frac12$. Then, the rest of the weights is repartitioned between PopTrack and t-CoMem, with an empirically-motivated preference for t-CoMem (see the results of individual components in Figure~\ref{fullMRR}). For $w_{\text{ not}}$, the inverse logic follows: since EdgeBank is not confident, it should not be the principal component, and given t-CoMem's empirical superiority, we set that third component to be the principal one -- thus achieving a weight of $\frac12$. Likewise, the rest of the weights is shared between EdgeBank and PopTrack, with a preference for PopTrack, given the lack of confidence in EdgeBank's score. 

\vskip .1in 

\texttt{Multi\_conf(eb\_score, pop\_score)} follows a similar logic, while incorporating two confidence signals: the existing \texttt{eb\_score}, as well as \texttt{pt\_score}, it's PopTrack analog. Recalling PopTrack's logic, it gives a score of 1.0 if the destination node being inquired is in the top-$K$ most popular nodes, and 0 otherwise. This score is the second confidence signal \texttt{multi\_conf} relies on. Similarly to \texttt{EB\_conf}, the model assigns different fixed weights depending on the confidence flags. We now have four cases, as presented in Table~\ref{tab:cases}, which are based on heuristic conditional weighting that reflects the confidence signals. When both signals are reliable, EdgeBank and PopTrack are given the highest weights, while t-CoMem gets 20\%. When only one of the two signals is positive, the other is downweighted: if EdgeBank is stronger, it gets 45\%, while, if PopTrack is, it is given 70\%. This difference is partially empirically motivated, and partially due to the consideration that t-CoMem relies on PopTrack (and thus, that strongly weighting both could be redundant). Finally, in the case where both signals are unreliable, t-CoMem is given more importance than EdgeBank. Generally, this scheme prioritizes the PopTrack weight. This choice is a consideration of PopTrack's strong performance in high-surprise datasets, most notably \texttt{tgbl-review}. Considering t-CoMem's high performance in low-surprise settings, it becomes important to combine these strengths. 

\begin{table}[h!]
\centering
\begin{tabular}{c|ccccc}
\toprule
\textbf{Case} & \texttt{eb\_score} & \texttt{pop\_score} & $\alpha$ & $\beta$ & $\delta$ \\
\hline
1 & 1 & 1 & 0.35 & 0.45 & 0.20 \\
2 & 1 & 0 & 0.45 & 0.25 & 0.30 \\
3 & 0 & 1 & 0.15 & 0.70 & 0.15 \\
4 & 0 & 0 & 0.20 & 0.45 & 0.35 \\
\bottomrule
\end{tabular}
\caption{Weight configurations in \texttt{multi\_conf}}
\label{tab:cases}
\end{table}

\end{document}